# Developing an ICU scoring system with interaction terms using a genetic algorithm[•, ♦]

**Chee Chun, Gan**[a,*], **Gerard P., Learmonth**[b]

[a]**Department of Systems and Information Engineering, University of Virginia, Charlottesville, VA, USA**
[b]**Frank Batten School of Leadership and Public Policy**

## Abstract

ICU mortality scoring systems attempt to predict patient mortality using predictive models with various clinical predictors. Examples of such systems are APACHE, SAPS and MPM. However, most such scoring systems do not actively look for and include interaction terms, despite physicians intuitively taking such interactions into account when making a diagnosis. One barrier to including such terms in predictive models is the difficulty of using most variable selection methods in high-dimensional datasets.

A genetic algorithm framework for variable selection with logistic regression models is used to search for two-way interaction terms in a clinical dataset of adult ICU patients, with separate models being built for each category of diagnosis upon admittance to the ICU. The models had good discrimination across all categories, with a weighted average AUC of 0.84 (>0.90 for several categories) and the genetic algorithm was able to find several significant interaction terms, which may be able to provide greater insight into mortality prediction for health practitioners. The GA selected models had improved performance against stepwise selection and random forest models, and provides greater flexibility in terms of variable selection by being able to optimize over any modeler-defined model performance metric instead of a specific variable importance metric.

---

[•] Conflicts of interest: We have no competing interests.

[♦] Role of the funding source: We received no funding for this study.

[*] Corresponding author: Department of Systems and Information Engineering, University of Virginia, 151 Engineer's Way, P.O. Box 400747, Charlottesville, VA 22904, Tel.: 434-924-5393
E-mail address: cg8pa@virginia.edu (Chee Chun Gan)

# 1 Introduction

Predictive modelling in healthcare is a rapidly growing field. Recent innovations in information systems use in hospitals has resulted in a massive increase in the availability and accuracy of patient electronic health records (EHRs) and other sources of medical data. This Data Science boom has enabled the development of more predictive analytics tools to aid health practitioners in tasks such as diagnosing illnesses, assessing the likelihood of patient readmission, and predicting patient mortality.

Many predictive scoring systems for adult ICU patient mortality have been developed. Among the most popular are the Acute Physiology and Chronic Health Evaluation (APACHE) score, the Mortality Probability Models (MPM) and the Simplified Acute Physiology Score (SAPS). Most of these predictive models are built using physiological, clinical or therapeutic variables that are routinely collected in the ICU, either as a first day snapshot or dynamically updated throughout a patient's ICU stay. Furthermore, most such scoring systems are based on a form of logistic regression to predict a patient's probability of mortality.

The Acute Physiology And Chronic Health Evaluation (APACHE) score was developed by Knaus et. al. [1] to assess the severity of illness of critically ill adult patients admitted to the intensive care unit (ICU). The first APACHE model consisted of 34 physiologic predictors selected using expert judgement. Further refinements to the APACHE model have followed with APACHE II, III and IV. APACHE II has been widely used in many hospitals and healthcare facilities for benchmarking purposes [2].

The APACHE II score is based on several clinical and physiologic measurements taken when a patient is first admitted to the ICU [3]. For APACHE II, the score is calculated from the following 13 predictors: age, alveolar-arterial gradient (A-aO$_2$) or partial pressure arterial oxygen (PaO$_2$) depending on the fraction of inspired oxygen (FiO$_2$), rectal temperature, mean arterial pressure (MAP), arterial pH, heart rate, respiratory rate, sodium (serum), potassium (serum), creatinine, hematocrit, white blood cell count and Glasgow Coma Scale score. The predictor values are used in a univariate logistic regression model to predict mortality. APACHE III expanded on APACHE II by including five additional physiologic predictors to the APS component, and included three two-way interaction terms as well [4]. The latest version, APACHE IV, uses a multivariate logistic regression model with a much larger dataset (110,588 patients) compared to its predecessors [5]. However, the exact variables and methodology used in APACHE IV have not been published.

The Simplified Acute Physiology Score (SAPS) was originally based on the APS predictors included in APACHE [6]. Expert judgement was used to reduce the number of predictors to 13 physiologic variables and patient age. SAPS II later included 4 additional demographic variables, bringing the total up to 17 (12 physiologic, 5 demographic) [7]. The predictor values were assigned a score by binning over the range of values (similar to APACHE), with the sum of these scores then being used in a logistic regression for patient mortality. The Mortality Probability Model (MPM) scoring system was developed using 12 variables in a multivariate logistic regression model [8]. Initially based only on data at the time of admission, further studies incorporated data taken 24 hours and 48 hours after admission. MPM II was later developed which included models built for data at admission, after 24 hours, after 48 hours and after 72 hours [9]. Two-way interaction terms were considered in MPM II, but were eventually rejected for not satisfying the author's criteria for inclusion.

The aforementioned ICU scoring systems have been validated with good performance in numerous studies [10]. However, a common point among these scoring systems is that they are mainly developed using predictors selected by subject matter expert judgement and mostly do not include interaction terms (APACHE III and possibly APACHE IV are exceptions). However, intuitively when predicting patient mortality it is likely that the existence of certain conditions in conjunction may pose a much greater health risk than when these conditions exist independently. Many physicians would naturally take into account the interplay of all physiologic variables when making a diagnosis, instead of considering each variable independently. In a complex problem such as predicting mortality, there may be many interaction effects that can give additional power to the model. In many cases, health practitioners are aware of such effects based on their experience and judgement but have no way of quantifying the strength of the interactions due to the lack of research into the inclusion of interaction terms. Thus, for the sake of model parsimony interaction terms are often omitted (e.g. the MPM II model rejected interaction terms if there was no "clinical plausibility" behind them [9]). However, it is also possible that beneficial interaction effects exist which are currently unknown to health practitioners and therefore would not be included in a model designed mainly using expert knowledge.

Thus, this exploratory study aims to develop a prototype ICU mortality scoring system using machine learning methods (a genetic algorithm) for variable selection instead of relying solely on expert knowledge. By evaluating the efficacy of models with interaction terms included, we aim to explore the potential benefits of using a variable selection method that can handle a large number of interactions to develop such models and hopefully find novel interactions that may not be well-known to health practitioners.

## 2    ICU mortality dataset

For this study, we obtained a dataset of 224,418 patient records with 12 binary comorbidities, 5 categorical clinical predictors, and 2 numeric predictors[1]. A similar dataset was used to evaluate APACHE IV against APACHE III [5]. Table 2.1 below summarizes the list of predictors included in the dataset.

Table 2.1:  Predictors in ICU dataset

| Binary predictors | Categorical predictors | Numeric predictors |
|---|---|---|
| operative, emergency, aids, myeloma, lymphoma, cirrhosis, tumorwm, immunosup, hepfail, copd, diabetic, dialysis | visit priorloc gender ethnic dx_group | age APS |

---

[1] Private communication with Dr. Andrew Kramer, formerly of the Cerner Corporation.

The first two binary predictors represent whether a patient is in the ICU for an operative or emergency procedure. The remaining binary predictors represent the absence or presence (0 or 1 respectively) of the listed comorbidities in the patient upon being admitted to the ICU. For the numeric predictors, "age" lists the patient's age in years (integer) while the Acute Physiology Score (APS) is an integer score based on a regression model using 12 clinical predictors, some of which are included in the list of APACHE II predictors.

The first categorical predictor, "visit", indicates how many times the patient has been admitted to the ICU and ranges from 1 to 9. "Priorloc" indicates the patient's location prior to entering the ICU, e.g. home, other hospital ICU etc. "Gender" and "ethnic" indicate the sex and ethnicity (6 levels) of the patient respectively. "Dx_group" stores the patient's diagnosis code, which is given by a physician upon admittance to the ICU. The diagnosis code is assigned based on the physician's diagnosis of the patient's condition. The diagnosis code is a factor (with 122 levels in this dataset) which can be grouped into 16 categories. Note that a patient being admitted to the ICU can exhibit multiple conditions, e.g. head trauma and intercerebral hemorrhage. However, only a single primary condition (as judged by the attending physician) is recorded in the data. Thus, each patient can only be associated with a single "dx_group" value.

While the ICU dataset was closely related to the data used to develop the APACHE models, several key variables were omitted (Glasgow Coma Scale, AaDO2/PaO2, pH arterial, potassium). Thus, we were unable to calculate the APACHE II score (or any of the other commonly used ICU scoring systems) for the patients in the dataset as a baseline comparison.

## 3      Data preprocessing

The dataset included two binary predictands, "icudead" and "hosdead". These labels represent whether the patient passed away in the ICU or subsequently in the hospital after being discharged from the ICU. For the purposes of this study, we only considered patient mortality in the ICU as there could be multiple complicating factors involved in hospital mortality that are not captured in the dataset. Thus, all patients who passed away in the hospital were removed, leaving only patients that survived or passed away in the ICU. Records that contain missing data in the categorical/binary predictors were removed, while records with missing data in the numeric predictors were replaced by the mean. As a result, the final dataset used for the analysis consisted of 154,281 patient records.

Several issues arose during the initial analysis of the dataset. Firstly, the APS predictor is an aggregate measure of several clinical predictors. Thus, it provides a general indication of the patient's health condition but does not provide information on the factors contributing to the score. While it performed adequately as an input to the original APACHE formulation, in order to explore potential interaction terms (especially with comorbidities) it would be more meaningful to expand the APS into its constituent components. Doing so added 12 additional numeric predictors to the dataset. These numeric predictors, together with the age of the patient, were scaled to have mean 0 and variance 1 in order to reduce the effects of multi-collinearity.

Secondly, the "visit" variable was changed from a 9 level factor to a 2 level factor indicating whether the patient was a first time visitor to the ICU, or a repeat visitor. Repeated visits to the ICU could be indicative of additional health complications or a poor health condition in general, leading to higher risk of mortality. However, the vast majority of patients had "visit" levels of either 1 or 2 (>98%), with a small minority having more than 2 visits. As such, we could combine all "visit" levels 2 or greater into a single level, greatly reducing the complexity of the model while retaining most of the predictive power.

Lastly, the "dx_group" predictor with 122 levels resulted in a very sparse matrix with many diagnosis codes belonging to very few patients, or none at all. In addition, consultations with subject matter experts (physicians at the University of Virginia Hospital ICU) revealed that in many cases the initial diagnosis is subjective and the diagnosis code assigned to the patient can vary substantially from physician to physician. Thus, the existing data on diagnosis codes is likely to be fairly noisy. However, there is less contention regarding the category of diagnosis. For example, it may be unclear whether a patient is suffering from bacterial pneumonia or viral pneumonia, but most physicians would categorize the diagnosis as a respiratory condition.

Following this line of reasoning, the various diagnosis codes were aggregated into the following 12 categories shown in Table 3.1.

Table 3.1: Diagnosis categories in ICU dataset

| Category | # of patients |
|---|---|
| Cardiovascular diagnosis | 52,630 |
| Cardiovascular surgery | 9,690 |
| Respiratory diagnosis | 23,047 |
| Respiratory surgery | 3,478 |
| Neurologic diagnosis | 20,222 |
| Neurologic surgery | 6,510 |
| Gastrointestinal diagnosis | 11,422 |
| Gastrointestinal surgery | 8,975 |
| Trauma diagnosis | 6,869 |
| Trauma surgery | 2,261 |
| Metabolic diagnosis | 6,839 |
| Genitourinary diagnosis | 2,338 |
| ***Total*** | ***154,281*** |

Furthermore, our discussion with subject matter experts suggested that it would likely improve model performance to subset the data according to the categories shown above. A patient admitted to the ICU for trauma injuries could have a very different set of mortality predictors than a patient admitted for respiratory problems. Many of the original ICU scoring systems were intentionally designed for ease of use with pen and paper calculations, and developing different models for different diagnosis codes would have greatly complicated the scoring process. However, with the widespread use of information technology in hospitals it should no longer be a requirement to be constrained to a single aggregated model for all patient conditions. By developing a model for each diagnosis category, we are also able to better explore potential interaction terms without the confounding effects of other conditions. Appendix A lists the final predictors considered in the models.

For each subset, logistic regression models were used to fit the data to predict ICU mortality. This choice was informed by several factors. Firstly, logistic regression models are widely used and well-understood by physicians. Predictions from logistic regression models are also easier to calculate without special software and can be performed using spreadsheets or mobile apps, compared to models such as random forests or artificial neural networks. Secondly, many studies of ICU mortality have used logistic regression models with similar predictors and demonstrated good performance. One of the primary concerns with logistic regression models is the possible presence of non-linear predictors, which are common in medicine due to the prevalence of homeostatic processes in living organisms. However, empirical results show that logistic regression models perform well on many medical datasets even without first applying transformations to non-linear predictors. Lastly, logistic regression models are easily interpretable, especially with regards to interaction terms. Interaction terms are explicitly defined in logistic regression models and thus their effects can be more easily isolated and evaluated.

After deciding on the model, we now have to determine the appropriate quantitative metric to use for model evaluation. As the GA provides great flexibility in the choice of fitness function, there are many possible options. The area under the Receiver Operating Characteristic (ROC) curve is a metric that is commonly used in machine learning for model comparison and has also seen widespread use and acceptance in the medical community. The ROC curve is derived by using the model's predictions to plot the true positive rate (TPR) against the false positive rate (FPR) for various values of the decision threshold. The area under the ROC curve (AUC) can therefore be used as a metric of a model's discriminative power, with a larger AUC indicating that a model has a higher probability of ranking a randomly chosen positive instance higher than a randomly chosen negative instance. It should be noted that the AUC alone should not be taken as a definitive measure of a model's effectiveness. A model with a higher AUC does not necessarily perform better than another model with a lower AUC, as the AUC represents the models' performance across all possible thresholds. When a model is used for classification a specific threshold has to be chosen in order for a class prediction to be made, and the relative performance of the models at that specific threshold could well differ from their AUC rankings. Nevertheless, we chose to use AUC as the model evaluation criteria as the AUC serves well as a general indicator of model performance and has been used extensively in evaluating APACHE, SAPS, MPM and other such scoring systems.

# 4 Using a genetic algorithm for variable selection

A genetic algorithm was developed to select potential main effect and interaction terms for the logistic regression model. The genetic algorithm consists of a population of candidate predictor sets, termed "chromosomes", which are evaluated according to a user-specified fitness function at each generation of the algorithm. High performing chromosomes have a higher chance to pass on to the next generation, with some modifications such as recombination with other high performing chromosomes or random mutation (performed according to a varying probability parameter). A maximum chromosome size is defined to set an upper bound for the maximum number of predictors included in the final solution obtained at the end of the specified number of generations. Appendix B describes the GA selection procedure in greater detail.

For each diagnosis subset of the ICU dataset, the GA framework was used to perform variable selection for a logistic regression model, using AUC as the fitness function. Each subset was split into ten folds for cross-validation using random sampling without replacement, with the size of folds 1-9 being set to floor($\frac{N}{10}$), where N is the total number of records in the subset. Fold 10 contains the remaining records after folds 1-9 have been drawn. Each candidate solution consists of a set of predictors, which are then evaluated on each test fold in turn after being trained on the remaining nine folds, ensuring that the test data never overlaps with the training data. The ten resulting AUC scores are then averaged to obtain the overall AUC score for the aforementioned set of predictors. Table 4.1 below shows the settings for the GA's meta-parameters used for all subsets.

Table 4.1: GA parameter settings

| | |
|---|---:|
| Chromosome population size | 30 |
| Min/Max number of predictors | 5/100 |
| Maximum number of generations | 250 |
| Recombination probability | 0.5 to 0.2 |
| Mutation probability | 0.01 to 0.2 |

For each diagnosis subset, 5 runs were performed using different initial random number generator seeds, and the best performing GA result was chosen. The following section describes the results from each subset, as well as provides some comparisons with other modelling methods.

## 5     Results and discussion

Due to the limitations of the variables provided in the dataset, we were unable to compare the AUC of the GA selected model against other ICU scoring systems like APACHE, SAPS II and MPM II. To provide a comparison, for each subset we developed a logistic regression model using stepwise selection according to Akaike Information Criterion (AIC) and a random forest model with 500 trees. The same ten folds used for the GA selection process were used to evaluate the AUC with each of the stepwise selected logistic regression models and random forest models. Table 5.1 below shows the mean AUCs obtained for the stepwise-selected logistic regression model, the random forest model, and the GA-selected logistic regression model respectively. For each subset, the standardized mortality ratio (SMR) of the GA-selected logistic regression models was not significantly different from 1.0, indicating no major differences between the observed number of deaths and the expected number of deaths. The Wilcoxon signed rank test was used to test for significance of the difference in AUCs between the GA vs stepwise selection and GA vs random forest by comparing the paired AUC scores in each fold used in cross-validation for the two selection methods evaluated.

Table 5.1: Mean AUC for logistic regression (stepwise), random forest and logistic regression (GA)

|  | Stepwise | Random forest | GA | GA vs Step | GA vs RF |
|---|---|---|---|---|---|
|  | *AUC* | *AUC* | *AUC* | *p-vals\** | *p-vals\** |
| **Cardiovascular diagnosis** | 0.8187 | 0.8605 | 0.8300 | 0.0039 | 0.0020 |
| **Cardiovascular surgery** | 0.8614 | 0.8684 | 0.8921 | 0.0137 | 0.0840 |
| **Respiratory diagnosis** | 0.7719 | 0.7761 | 0.7852 | 0.0020 | 0.1602 |
| **Respiratory surgery** | 0.8290 | 0.8213 | 0.9159 | 0.0098 | 0.0059 |
| **Neurologic diagnosis** | 0.7824 | 0.8390 | 0.8050 | 0.0137 | 0.0020 |
| **Neurologic surgery** | 0.8833 | 0.8678 | 0.9200 | 0.0020 | 0.0020 |
| **Gastrointestinal diagnosis** | 0.8265 | 0.8383 | 0.8426 | 0.0039 | 0.4922 |
| **Gastrointestinal surgery** | 0.8199 | 0.8545 | 0.8692 | 0.0137 | 0.0020 |
| **Trauma diagnosis** | 0.8170 | 0.8805 | 0.8597 | 0.0039 | 0.0840 |
| **Trauma surgery** | 0.8383 | 0.8896 | 0.9065 | 0.0039 | 0.1934 |
| **Metabolic diagnosis** | 0.8560 | 0.8580 | 0.8952 | 0.0020 | 0.0098 |
| **Genitourinary diagnosis** | 0.7844 | 0.7855 | 0.8599 | 0.0020 | 0.0273 |

*\* p-values calculated by performing Wilcoxon signed-rank test on AUC scores for cross-validation folds*

It can be seen that the discrimination of the GA-selected model is fairly good, ranging from 0.7852 to 0.9200 across the various subsets. The GA-selected model significantly outperformed the stepwise selected model in all 12 categories (at a 0.05 significance level), while the random forest model was better in 2 categories and worse in 5 categories. In particular, the GA-selected model performed markedly better than both stepwise selection and random forest in the "respiratory surgery" and "genitourinary diagnosis" categories.

Tables 5.2 and 5.3 below summarize the significant predictors in each category. Each column represents a diagnosis category, while the rows represent the main effects terms. A highlighted cell in a column indicates that the main effect is included in the model for the indicated diagnosis category. The numbers in each cell indicate the variables with which the term has significant pair-wise interactions. For example, in the model for cardiovascular diagnosis the "visit" term is a significant main effect with no interaction terms, while the "age" term has significant interactions with "dialysis", "temp", "sodium" and "album".

Table 5.2 : Model summary for diagnosis categories 1-6

|   |   | Cardio diag | Cardio surg | Resp diag | Resp surg | Neuro diag | Neuro surg |
|---|---|---|---|---|---|---|---|
| 1 | visit |  | 4,20 | 20,28 |  |  | 26 |
| 2 | ipriorloc | 12,18 |  | 11,14 |  | 4,16,21 |  |
| 3 | gender |  |  |  |  |  | 23 |
| 4 | age | 16,18,25,28 | 1,26 | 11 |  | 2,22,24 |  |
| 5 | operative |  |  |  |  |  |  |
| 6 | emerg |  | 29 |  |  |  | 19 |
| 7 | aids |  |  |  |  |  |  |
| 8 | myeloma | 25 |  |  |  | 12,23 |  |
| 9 | lymphoma |  |  |  |  |  |  |
| 10 | cirrhosis |  |  |  |  | 24 |  |
| 11 | tumorwm |  | 18 | 2,4,12 |  |  |  |
| 12 | imm.sup | 2 |  | 11 |  | 8,28 |  |
| 13 | hepfail | 26,27 |  | 28 |  | 25 |  |
| 14 | copd | 19,22 |  | 2,26 |  | 22 |  |
| 15 | diabetic | 22,27 | 16 |  | 21 | 25,27 | 22 |
| 16 | dialysis | 4,26 | 15,21,26 | 26 |  | 2 |  |
| 17 | ethnic | 18,25 | 24 |  |  | 19 |  |
| 18 | temp | 2,4,19,20,27 | 11,28 | 19 | 20 | 20,22,26,28 | 19,22,29 |
| 19 | map | 20,21,22,25,27 | 20 | 18,20,21,23,26 |  | 17,21,28,29 | 6,18,23 |
| 20 | hr | 23 | 1,19,21 | 1,19 | 18,21 | 18 |  |
| 21 | rr | 25 | 29 | 19 | 15,20,22,25,27 | 2,19,22,23,27 |  |
| 22 | urine | 27 |  |  | 23 | 4,14,18,21,23,24,25 | 15,18 |
| 23 | wbc | 26,28 |  | 29 | 22 | 8,21,22,24 | 3,19 |
| 24 | hcrit | 26 | 17 |  |  | 4,10,22,23 | 25 |
| 25 | sodium | 4,8,17,19 |  |  |  | 13,15,22 |  |
| 26 | creat | 13,16,23,24 | 4,16 | 14,16,19 |  | 18,27 | 1 |
| 27 | gluc | 13,15,18,19,22 |  |  |  | 15,21,26,28 |  |
| 28 | album | 4,23 | 18 | 13,29 |  | 12,18,19,27 |  |
| 29 | bili |  | 6,21 | 23,28 |  |  | 18 |

Table 5.3 : Model summary for diagnosis categories 7-12

|   |   | Gastro diag | Gastro surg | Trauma diag | Trauma surg | Meta diag | Genito diag |
|---|---|---|---|---|---|---|---|
| 1 | visit | 15 |  |  |  | 24 |  |
| 2 | ipriorloc | 22,25 | 13 | 28 |  |  |  |
| 3 | gender |  | 18 | 18,23 | 26 | 28 | 22 |
| 4 | age | 22 |  | 22,25 | 19,21,27 | 11 |  |
| 5 | operative |  |  |  |  |  |  |
| 6 | emerg |  |  |  | 22 |  |  |
| 7 | aids |  |  |  |  |  |  |
| 8 | myeloma |  |  |  |  |  |  |
| 9 | lymphoma |  |  |  |  |  |  |
| 10 | cirrhosis |  |  |  |  |  | 15 |
| 11 | tumorwm | 22 |  | 22,24,27 |  | 4 |  |
| 12 | imm.sup | 23 |  |  |  |  | 27 |
| 13 | hepfail | 28,29 | 2,23 |  |  |  |  |
| 14 | copd |  |  | 15,23 | 26,28 |  |  |
| 15 | diabetic | 1,26 |  | 14 |  |  | 10 |
| 16 | dialysis | 24,25,26 | 22 | 20 | 20 |  | 27,28,29 |
| 17 | ethnic |  |  | 18 |  |  |  |
| 18 | temp |  | 3,22,24 | 3,17,19,21,23 | 26 | 24,26 |  |
| 19 | map | 20 | 23 | 18,23,25 | 4,23,26 | 20,22,28 | 20 |
| 20 | hr | 19,23 |  | 16,21,25,27 | 16,25,29 | 19 | 19 |
| 21 | rr |  |  | 18,20,25,29 | 4,25 |  | 22,28 |
| 22 | urine | 2,4,11 | 16,18 | 4,11 | 6 | 19,27 | 3,21 |
| 23 | wbc | 12,20 | 13,19 | 3,14,18,19,28 | 19 | 25,26 |  |
| 24 | hcrit | 16 | 18 | 11 |  | 1,18 |  |
| 25 | sodium | 2,16,27 |  | 4,19,20,21 | 20,21 | 23 |  |
| 26 | creat | 15,16,29 |  |  | 3,14,18,19 | 18,23 |  |
| 27 | gluc | 25 |  | 11,20,29 | 4 | 22 | 12,16 |
| 28 | album | 13 |  | 2,23 | 14 | 3,19 | 16,21 |
| 29 | bili | 13,26 |  | 21,27 | 20 |  | 16 |

The current implementation of the GA does not directly select for model sparsity in the fitness function, which results in the GA-selected models having a fairly large number of predictors. While a penalty for model size could be added to the fitness function, doing so comes with significant downsides. Firstly, such a penalty function could interfere with the GA selection process by forcing the GA to become too greedy and prematurely weed out predictors that may initially provide little improvement to the AUC, but would improve the fitness in the presence of certain other predictors. Secondly, the determination of the appropriate penalty function is non-trivial and has a significant effect on the GA's performance. However, it should be noted that the larger size does not necessarily translate to a larger burden on data collection, as the majority of the additional variables are interaction terms derived from the original 29 main effects variables which are already routinely collected. Furthermore, the GA-selected models can be further refined using expert judgement or other variable selection methods, both of which become more viable once the number of potential predictors has been reduced using the GA.

As expected, the models for each diagnosis category differ substantially. However, the patient's age, number of visits and various APS predictors generally are significant in almost every category, which is consistent with the findings of other ICU scoring systems. The presence of diabetes and whether the patient is on dialysis are also significant in several models, while the presence of AIDS, myeloma, cirrhosis, and whether the patient was admitted for operative purposes is relatively insignificant. Further examination of the models also reveals some interesting observations. Firstly, the "emergency" predictor is significant in most of the diagnosis categories pertaining to surgery. Secondly, most of the models include significant interactions amongst the APS predictors (together with significant APS main effects terms). These interaction effects have not been included in other ICU scoring systems that use the APS score as an aggregate predictor. Thirdly, the GA was able to identify several interactions in the dataset that could potentially be avenues for further study. For example, the ethnicity of the patient is significant in several categories (cardiovascular diagnosis, neurological diagnosis, trauma diagnosis) along with interactions with APS predictors such as sodium and temperature. The gender of the patient also has significant interactions with APS predictors in various diagnosis categories.

# 6 Conclusions

The results of the study show that there is potential benefit in utilizing machine learning methods, in this case a genetic algorithm for variable selection, in developing ICU scoring systems which include interaction terms. Using AUC as a measure of model performance, the GA-selected logistic regression models had comparable or better discrimination than stepwise-selected logistic regression models or random forest models. We also show that developing different models for various diagnosis categories rather than using a single model for all ICU patients may yield improved model performance as well as provide insight in the form of significant interaction terms for each particular diagnosis category. Thus, the GA selection process can serve as a useful first step in developing models to support physicians in predicting patient mortality.

However, the GA-selection procedure also comes with some notable drawbacks. The first is the procedure run-time, which can be very significant compared to other variable selection methods. On the other hand, the GA selection procedure is able to deal with an arbitrarily large number of potential predictors, unlike several common variable selection procedures. Furthermore, the long run-time is only applicable during model development (or updating) not during patient classification.

Secondly, there is no theoretical guarantee that the GA will find globally optimum models that generalize well. The models returned by the GA should be validated on another dataset that should ideally contain the same predictors used in other ICU scoring systems, which would allow a better comparison of the models with interaction effects. The GA could also be coupled with other variable selection procedures to try to prune the final number of predictors, which could make the models more generalizable to other datasets.

# Appendix A: List of predictors and descriptions

| Name | Description | Type |
|---|---|---|
| visit | # of times patient has been admitted to ICU | Factor (1-9) |
| ipriorloc | Prior location of patient | Factor (emergency department, other floor, home, ICU transfer, other hospital, other hospital ICU, other, SDU, telemetry) |
| gender | Male or female | Factor (0 = Male, 1 = Female) |
| age | Patient age in years | Numeric |
| operative | Procedure is operative | Binary |
| emerg | Procedure is emergency | Binary |
| aids | Presence of Acquired Immune Deficiency Syndrome (AIDS) | Binary |
| myeloma | Presence of myeloma (cancer of plasma cells) | Binary |
| lymphoma | Presence of lymphoma (cancer of lymphatic system) | Binary |
| cirrhosis | Presence of cirrhosis | Binary |
| tumorwm | Presence of tumor with metastasis | Binary |
| immunosup | Presence of immunosuppressive disorder | Binary |
| hepfail | Presence of hepatic failure | Binary |
| copd | Presence of chronic obstructive pulmonary disease | Binary |
| diabetic | Presence ofdiabetes | Binary |
| dialysis | Patient is on dialysis | Binary |
| ethnic | Ethnicity of patient | Factor (other unknown, African American, Asian, Caucasian, Hispanic, Native American) |
| temp | Temperature | Numeric |
| map | Mean arterial pressure | Numeric |
| hr | Heart rate | Numeric |
| rr | Respiratory rate | Numeric |
| urine | Urine output | Numeric |
| wbc | White blood cell count | Numeric |
| hcrit | Hematocrit | Numeric |
| sodium | Sodium level | Numeric |
| creat | Creatinine level | Numeric |
| gluc | Glucose level | Numeric |
| album | Albumin level | Numeric |
| bili | Bilirubin level | Numeric |

# Appendix B: Genetic algorithm modification for including interaction terms in high-dimensional datasets

The biggest challenge in the inclusion of interaction terms in variable selection problems is the dramatic increase in the solution space. For now, we constrain ourselves to only considering second order interaction terms, i.e. only pair-wise interactions. For n main effects terms, this adds $\binom{n}{2}$ second order interaction terms. For relatively small n the additional terms can still be handled using the traditional GA variable selection chromosome (a single vector of 0-1 bits of length n+$\binom{n}{2}$ to indicate all possible variables), but this implementation quickly becomes unwieldy. For 100 variables an additional 4,950 interaction terms are added, and for 200 variables this jumps to 19,900. Thus for problems with hundreds of variables a new chromosome formulation is needed.

In order to solve the scalability issue, we propose some modifications to the original chromosome formulation. While only second order interaction terms are examined here, the basic technique for extending the GA framework remains applicable for higher order interactions at the cost of greatly increased computation time. Firstly, a maximum chromosome length l is defined. This allows the modeler to specify an upper bound for model sparsity, as in many instances modelers may not be interested in creating a model with thousands of variables. Secondly, instead of each bit in the chromosome simply being 0-1 to indicate the absence or presence of a variable, each bit now stores the index number of a variable to be included, and 0 if the bit is a "dummy bit". Dummy bits are placeholder bits within the chromosome that reserve space for a potential variable to enter the model. This formulation allows for chromosomes representing models with a differing number of included variables while still allowing chromosome length to be homogenous within the population, which simplifies the crossover operation.

Figure B.1: Chromosome with dummy bits

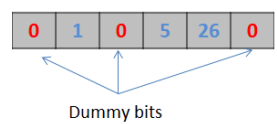

The chromosome in Figure B.1 shows a chromosome of length 6 with 3 dummy bits, with variables 1,5 and 26 included in the model. Each new chromosome is initialized with dummy bits in all positions, and the number of initial variables is chosen uniformly between 1 and L (maximum number of variables). Pre-seeded variables can also be utilized instead of random selection. The index positions of these variables are also chosen by sampling without replacement from the available L positions, after which the variables (either randomly chosen or pre-seeded) are then filled into their respective index positions on the chromosome.

After determining the fitness levels of all members of the population, a selection procedure is then used to choose several parent chromosomes. One common selection method is tournament selection, where several candidates are chosen randomly to participate in a "tournament" during which the fitness values of competing chromosomes are compared, with the winner being selected as a parent chromosome. This parallels the biological process of natural selection where more fit individuals in a population have a greater chance of reproducing and passing on their genes to their offspring. In our implementation, the chromosome with the maximum fitness is automatically passed on to the next generation (elitist selection to preserve the best found solution), along with a mutation of the chromosome with the maximum fitness. The remaining members of the population are chosen by tournament selection, with each tournament containing 10 randomly selected candidates.

Once parent chromosomes have been selected, the crossover operation is used to generate offspring, or child chromosomes. Again, there are various forms of crossover operators used with the underlying notion of combining the genes from multiple (usually two) parent chromosomes into a single offspring. The most basic crossover operator is a fixed point crossover, with the crossover point usually being the midpoint of the chromosome. The underlying notion behind the crossover operator is that a high-performing parent chromosome should contain certain elements that contribute to its fitness score. In the case of a variable selection problem, it could be that high performing chromosomes contain a larger ratio of the "correct" variables. By combining the chromosomes of two parents, the crossover operator attempts to generate children which also have a high likelihood of equal or improved performance. The crossover operator can be applied according to a predefined probabilistic parameter setting. For example, a crossover probability of 0.5 would indicate that a pair of parents would have their chromosomes combined half the time. The other half of the time would see both parents being passed on to the next generation without mixing their chromosomes, similar to elitist selection.

The current chromosome formulation can handle an arbitrary number of main effects terms in addition to interaction terms as long as the modeler specifies a maximum number of variables. As the chromosome length is homogenous, the aforementioned single point crossover operator can be applied on the modified chromosome, with some additional checks to ensure that duplicate variables are removed. The mutation operator is separated into two types, a deletion mutation and an addition mutation. The deletion mutation replaces a random non-dummy bit with a value of 0, converting it to a dummy bit and removing the selected variable from the model. The addition mutation replaces a random dummy bit with a randomly selected variable that is currently not included in the model. Both types of mutation occur independently with probabilities $P_a$ and $P_d$ specified by the modeler. Both mutations occur simultaneously with probability $P_a * P_d$, resulting in one variable being switched out for another.

Both crossover and mutation (addition and deletion) are applied with a variable probability across the GA's run time. A minimum and maximum probability is defined for each operator (the same parameters apply to both types of mutation). The probabilities for each operator are adjusted each generation so as to vary from minimum to maximum or vice versa. The crossover probability $p_c$ is initialized to $p_{c\_max}$ = 0.5 at generation 0, and then varies across each generation i according to Equation 1 below until finally reaching $p_{c\_min}$ = 0.2 after *maxgen* iterations.

$$p_c(i) = p_{c\_max} - \left(\frac{i}{maxgen}\right)(p_{c\_max} - p_{c\_min}) \qquad (1)$$

The mutation probability $p_m$ (for both addition and deletion) is initialized to $p_{m\_min}$ = 0.01 and varies linearly throughout the run until it reaches $p_{m\_max}$ = 0.2 after *maxgen* iterations, as shown in Equation 2.

$$p_m(i) = p_{m\_min} + \left(\frac{i}{maxgen}\right)(p_{m\_max} - p_{m\_min}) \tag{2}$$

These varying probabilities are chosen to obtain a higher chance of crossover with less mutation at the beginning of the GA run (increased exploration of solution space), and a lower chance of crossover with more mutations at the end of the run (increased exploitation of good solutions in population).

In addition, the GA framework ensures the model obeys strong hierarchy. Each time an interaction term enters the model through either recombination or the addition mutation, a check has to be performed to ensure that the corresponding main effects terms are also included. If not, the missing main effects terms are inserted into random dummy bit positions. If a main effect term is deleted through the deletion mutation, then all interaction terms that include the aforementioned main effect term are also deleted.

Lastly, in order to prevent selection of models that over-fit the data, all fitness functions are evaluated using 10-fold cross-validation. The data is partitioned into ten folds, with the models being successively tested on a single fold and trained on the other nine folds. The final fitness is then obtained by averaging the model fitness over all ten test folds. With this process, there is never any overlap between data used for training models, and data used for evaluating the fitness.